\documentclass{article}

\usepackage{microtype}
\usepackage{graphicx}
\usepackage{subfigure}
\usepackage{booktabs} 

\usepackage{amsmath}
\usepackage{amssymb}
\usepackage{natbib}
\usepackage{adjustbox}

\usepackage{hyperref}

\usepackage[accepted]{icml2018}

\icmltitlerunning{Direct Preference Optimization for Chatbot Fine-Tuning}

\begin{document}

\twocolumn[
\icmltitle{Direct Preference Optimization for Chatbot Fine-Tuning: An Empirical Study}



\icmlsetsymbol{equal}{*}

\begin{icmlauthorlist}
\icmlauthor{Dezhi Yu}{equal,to}
\icmlauthor{Yvonne Qiu}{equal,to}
\icmlauthor{ShuoJia Fu}{equal,to}

\end{icmlauthorlist}

\icmlkeywords{Machine Learning, ICML}

\vskip 0.3in
]



\begingroup
\renewcommand{\thefootnote}{*}
\footnotetext{These authors contributed equally to this work.}
\endgroup

\begin{abstract}
We present an approach to fine-tuning large language models using Direct Preference Optimization (DPO), a reinforcement learning technique. Our experimental results demonstrate that DPO simplifies the training pipeline, improves computational efficiency, and achieves competitive performance. The evaluation using BLEU, ROUGE, and cosine similarity metrics indicates effective learning and convergence, though further investigation is needed to address observed training instability.

\end{abstract}

\section{Introduction}

The rapid advancement of large language models (LLMs) has revolutionized various natural language processing (NLP) tasks, ranging from machine translation and text summarization to dialogue generation and sentiment analysis. Traditional fine-tuning methods for these models often involve extensive computational resources and intricate training pipelines that include intermediate steps like reward modeling. To address these challenges, we explore Direct Preference Optimization (DPO), a reinforcement learning approach that directly optimizes the model based on human preferences without the need for reward modeling. The motivation of using DPO is because It is an algorithm that effectively optimizes the same objective as traditional RLHF (Reinforcement Learning with Human Feedback) algorithms, which is reward maximization constrained by KL-divergence. However, DPO stands out for its simplicity in implementation and ease of training. Essentially, the DPO update mechanism enhances the relative log probability of preferred responses over non-preferred ones, incorporating a dynamic, per-example importance weight. This approach mitigates the model degeneration often seen with a naive probability ratio objective. Similar to existing algorithms, DPO relies on a theoretical preference model, such as the Bradley-Terry model, to measure how well a reward function aligns with empirical preference data. Unlike traditional methods that use this preference model to train a reward model and subsequently optimize a policy based on the learned reward model, DPO redefines the preference loss directly in terms of the policy through a change of variables. This allows DPO to optimize a policy using a straightforward binary cross-entropy objective based on a dataset of human preferences, ultimately producing an optimal policy aligned with an implicit reward function derived from the preference data.\citep{Rafailov2023}.

In this report, we detail our approach to fine-tuning a pre-trained language model, cognitivecomputations/dolphin-2.1-mistral-7b, using DPO. We leverage a high-quality, human-labeled preference dataset to train the model, aiming to achieve efficient and effective performance across various NLP tasks. Our experimental results showcase the advantages of DPO in terms of simplification, efficiency, and adaptability, while also highlighting areas for further improvement.

Our study builds upon these foundations by applying DPO to a pre-trained LLM, cognitivecomputations/dolphin-2.1-mistral-7b, and evaluating its performance on various NLP tasks. The DPO framework we utilized is characterized by key equations that model the probability of one response being preferred over another and an optimization objective that maximizes this probability. The policy update rule is derived from the gradient of the loss function, ensuring efficient and effective learning.

To train our model, we used the argilla/ultrafeedback-binarized-preferences-cleaned dataset, which provides high-quality human-labeled preference data specifically designed for DPO training. This dataset addresses issues found in earlier datasets, such as mismatching labels, through a rigorous cleaning process. The dataset was randomly split into training and validation sets, with preprocessing steps to ensure efficient use of computational resources.

Our evaluation metrics include BLEU, ROUGE, and cosine similarity, which measure different aspects of text generation quality. BLEU focuses on precision, ROUGE emphasizes recall, and cosine similarity captures the semantic similarity between generated and reference texts. Additionally, we conducted evaluations using the WebGPT and Elizabot datasets to assess the model's performance in both knowledge-based and conversational contexts. Human evaluations further provided qualitative insights into the model's accuracy and conversational style.

In the following sections, we present our experimental results, detailing the training process, convergence of the loss function, and performance metrics. We discuss the implications of our findings and potential areas for further research, particularly addressing the observed instability during training and its possible causes.

\section{Background/Related Work}

The evolution of large language models (LLMs) has been marked by significant milestones, particularly with the advent of transformer-based architectures like GPT-3, BERT, and their successors. These models have set new benchmarks in numerous NLP tasks due to their ability to generate coherent and contextually relevant text. These chatbots can detect user intent, understand the context of the conversation, and provide personalized responses that feel more natural and tailored to the user’s needs. This level of contextual understanding and personalization is a significant step forward in the field of conversational AI, as it helps to bridge the gap between human and machine interactions.\cite{bert} However, fine-tuning these models for specific applications remains a complex and resource-intensive process. Traditional fine-tuning techniques often rely on supervised learning with labeled datasets, which can be limited by the quality and diversity of the training data.They also involve adjustments to all parameters, face challenges due to high computational and memory demands. This has led to the development of Parameter Efficient Fine-Tuning (PEFT) techniques, which selectively update parameters to balance computational efficiency with performance. \cite{balne2024parameter}

Recent advancements have introduced reinforcement learning (RL) approaches to address some of these limitations. Proximal Policy Optimization (PPO) and other RL algorithms have shown promise in optimizing LLMs by learning from interactions with the environment or human feedback. PPO, specifically, alternate between sampling data through interaction with the environment, and optimizing a “surrogate” objective function using stochastic gradient ascent whereas standard policy gradient methods perform one gradient update per data sample. \cite{schulman2017proximal}Despite these advancements, the process often involves an intermediate step of reward modeling, which adds to the complexity and computational burden.

Direct Preference Optimization (DPO) emerges as a novel approach that simplifies this process by directly optimizing the policy based on human preferences. Unlike traditional methods that require the creation and maintenance of a reward model, DPO leverages preference data to adjust the model's policy in a more straightforward manner. This approach not only reduces the computational overhead but also aligns the model more closely with human judgment and expectations.

\section{Approach }
We applied DPO to a pre-trained language model.

\subsection{Direct Preference Optimization (DPO)}
Direct Preference Optimization (DPO) is a reinforcement learning approach aimed at optimizing language models directly from human preferences without the intermediate step of reward modeling. Unlike traditional methods, DPO leverages a theoretical framework that allows the direct adjustment of the policy based on preference data \citep{Xu2024}.

The DPO algorithm can be described by the following key equations and concepts.

DPO uses a preference model where given a prompt $x$ and responses $y_1$ and $y_2$, the probability that $y_1$ is preferred over $y_2$ is modeled as:
\begin{equation}
P(y_1 \succ y_2 | x) = \sigma(r(x, y_1) - r(x, y_2)),
\end{equation}
where $\sigma$ is the sigmoid function and $r$ is the reward function \citep{Rafailov2023}.

The optimization objective of DPO is to maximize the following:

\begin{multline}
\mathcal{L}_{\text{DPO}} = \mathbb{E}_{(x, y_1, y_2) \sim \mathcal{D}} \left[ \log \sigma\left(\beta \left( \log \frac{\pi_\theta(y_1 | x)}{\pi_\theta(y_2 | x)} \right.\right.\right. \\
\left.\left.\left. - \log \frac{\pi_{\text{ref}}(y_1 | x)}{\pi_{\text{ref}}(y_2 | x)} \right) \right) \right]
\end{multline}
where $\beta$ is a constant scaling factor, and $\pi_\theta$ and $\pi_{\text{ref}}$ are the current and reference policies, respectively \citep{Xu2024, Rafailov2023}.

The policy update rule is derived from the gradient of the loss function:

\begin{align}
\nabla_\theta \mathcal{L}_{\text{DPO}} = \mathbb{E}_{(x, y_1, y_2) \sim \mathcal{D}} & \left[ \left( \sigma(\Delta r) - 0.5 \right) \right. \notag \\
& \left. \left( \frac{\nabla_\theta \pi_\theta(y_1 | x)}{\pi_\theta(y_1 | x)} - \frac{\nabla_\theta \pi_\theta(y_2 | x)}{\pi_\theta(y_2 | x)} \right) \right]
\end{align}

where $\Delta r = r(x, y_1) - r(x, y_2)$ \citep{Rafailov2023}.

DPO offers several advantages over traditional reinforcement learning methods:
\begin{itemize}
  \item \textbf{Simplification}: By directly optimizing the policy without learning an explicit reward model, DPO simplifies the training pipeline \citep{Xu2024}.
  \item \textbf{Efficiency}: The direct optimization process makes DPO computationally efficient, reducing the overhead associated with reward model learning and RL fine-tuning \citep{Rafailov2023}.
  \item \textbf{Performance}: Empirical studies have shown that DPO can achieve competitive or superior performance compared to PPO in various tasks, including sentiment modulation, summarization, and dialogue generation \citep{Xu2024, Rafailov2023}.
  \item \textbf{Adaptability}: DPO is robust to distribution shifts and can handle varying input distributions effectively \citep{Xu2024}.
\end{itemize}

\subsection{Pre-trained model}
The pre-trained model we utilize is \textbf{cognitivecomputations/dolphin-2.1-mistral-7b}, a fine-tuned version of the Mistral 7B model with a ChatML template. This model is based on the Mistral architecture by MistralAI and is distributed under the Apache-2.0 license. It has been trained on the Dolphin dataset, an open-source implementation of Microsoft's Orca. The model is uncensored, and the developers have filtered the dataset to remove bias and ensure alignment. Consequently, the model can respond to a wide range of questions, including potentially unethical ones, due to its high compliance with requests.

\subsection{Dataset}
\subsubsection{Description of Dataset}
The DPO Data source we use is \textbf{argilla/ultrafeedback-binarized-preferences-cleaned}, and we chose it because of the following reasons:
  \begin{itemize}
    \item This dataset provides high-quality, human-labeled preference data specifically designed for Direct Policy Optimization (DPO) training of large language models (LLMs).
    \item This dataset was created by Argilla to address issues found in earlier DPO datasets, such as mismatching labels.
    \item This dataset underwent a cleaning process that involved adding critiques and refining the existing data to ensure it accurately reflects real-world human preferences.
  \end{itemize}

\subsubsection{Training-validation spliting}
 We randomly split the dataset into 11000 for training, and 2750 for validation
 
\subsubsection{Preprocessing input}
During the training process, we set maximum prompt length to 512 and max length with combing prompt and answer to 1024, with truncation side left to reduce the memory requirement and computation cost by filtering out long samples and only keep the significant information. 

\subsection{DPO and Reference model structure}
\subsubsection {DPO Model structure}
We applied 4-bit precision quantization, PEFT with QLoRA configuration, and flash attention to enhance efficiency and scalability, especially for large-scale models and long sequences. For PEFT configuration, we used LoRa with lora alpha set to 128 and a dropout rate of 0.05. These three techniques significantly reduce the parameters and memory required for training and storage.
\subsubsection {Reference Model structure}
This model structure is used for the reference model.

\subsection{Loss Function}
We applied the \textbf{sigmoid function} as the loss function and treated the problem as \textbf{binary cross entropy loss}, with an \textbf{additional penalty of KL divergence}.The weight of KL divergence is controlled by beta, and the higher beta is, the less divergence we will get.

\subsection{Evaluation metrics}
\subsubsection{BLEU}
The BLEU (Bilingual Evaluation Understudy) score measures the correspondence between a machine-generated text and a reference text, focusing on precision. It is commonly used for evaluating the quality of text which has been machine-translated from one language to another.

\subsubsection{ROUGE}
The ROUGE (Recall-Oriented Understudy for Gisting Evaluation) score measures the overlap of n-grams, word sequences, and word pairs between the machine-generated text and the reference text, emphasizing recall.

\subsubsection{Cosine Similarity}
Cosine similarity is widely employed in text analysis and natural language processing applications to measure the similarity between documents or sentences. Each sentence is tokenized into a vector, and the cosine similarity is calculated as the dot product of these vectors divided by the product of their magnitudes. Unlike metrics such as BLEU and ROUGE, cosine similarity provides a more content-focused analysis, capturing the contextual and semantic similarity between texts more effectively.
\subsubsection {Human Evaluation}
We also manually reviewed the answers generated by the post-trained and pre-trained models, evaluating each response based on factual accuracy, coherence, overall usefulness and conversational style to determine which answer is better.

\subsubsection {Evaluation Methods}
  \begin{itemize}
    \item{\textbf{WebGPT dataset}}:
We utilized another dataset called WebGPT Comparisons, which contains prompts paired with two generated answers and their respective scores. For our analysis, we selected the first 100 prompts from this dataset and input them into both the post-trained and pre-trained models to generate responses. As a reference answer, we used the higher-scored answer from the WebGPT dataset. We then applied BLEU, ROUGE, and cosine similarity metrics to compare the answers generated by the post-trained model with the reference answers and the answers generated by the pre-trained model with the reference answers. Finally, we compared the average scores across the 100 prompts for both models.
\item {\textbf{Conversational Prompt}}:
\item We also provide the pre-trained and post-trained models with various conversational prompts (not knowledge-based questions) and record their responses. Then we employ a human labeling method to assess the performance of the models.\\\\The introduction of conversational prompts holds significance in this context due to the nature of WebGPT prompts. WebGPT primarily comprise knowledge-based questions with distinct correct answers. Consequently, these prompts fail to assess the speaking style of the response. Thus, by evaluating the performance of our post-trained model from conversational prompt, we aim to gain insights into its effectiveness across various language tasks.
\end{itemize}

\section{Experimental Results}
\subsubsection{Hyper-parameter}
we fine tune the hyperparameter and the final ones we applied are: beta:0.1; epoch: 1, batch size: 500, lr: 5*10e-5. 
\subsubsection{Training time and equip}
We have trained the model two times one with two A6000 GPUs with 13 hours and three ephoch, and the second time we changed to A100 with 4 hours and only 1 epoch. We'll present and analyze the results done by A6000 here, and results from A100 are added in the supplemental Materials section. 

\subsubsection{Training Loss Convergence}
\begin{figure}[ht]
\vskip 0.2in
\begin{center}
\centerline{\includegraphics[width=\columnwidth]{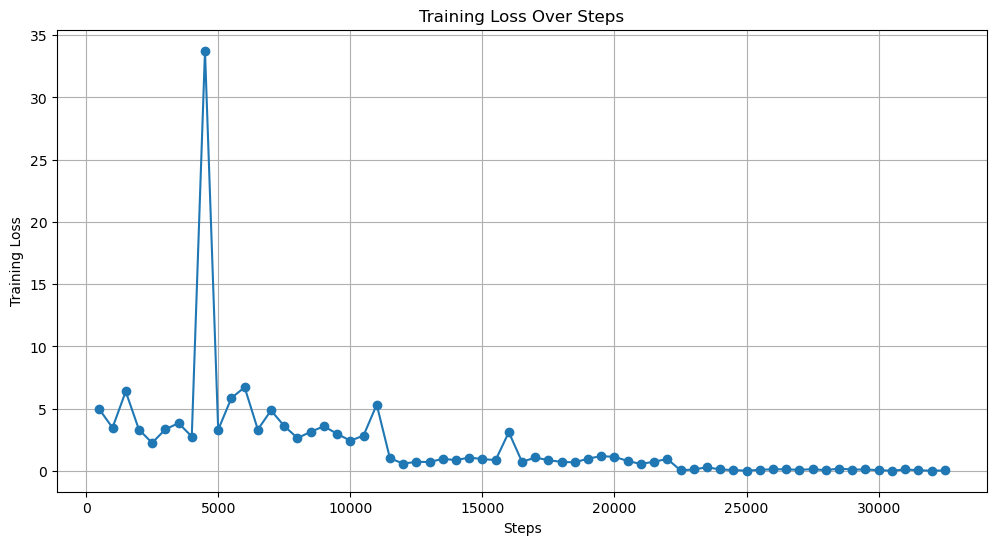}}
\caption{DPO Training Loss Over Steps}
\label{lossfig}
\end{center}
\vskip -0.2in
\end{figure}

The training starts with a relatively high loss (4.9535 at step 500) and exhibits significant variability in the initial stages. For example, the loss jumps to 6.3793 at step 1500 and then drops to 2.2520 by step 2500. There are instances of sudden spikes in the loss, such as at step 4500 where the loss jumps to 33.7498, indicating potential instability or outliers in the training process. This could be due to a number of factors including learning rate anomalies, data-related issues, or momentary GPU hardware issues. Despite the initial variability and occasional spikes, there is a general downward trend in the training loss as the steps progress. By step 11500, the loss drops significantly to 1.0191, showing an improvement in model performance. In the later stages of training, the loss values stabilize and show a consistent downward trend, reaching very low values like 0.0151 at step 30500. This indicates that the model is converging and learning effectively. Towards the end of the training process, there are a few very low loss values (e.g., 0.0483 at step 22500 and 0.0151 at step 30500), as the model have easier batches and benefitted from significant regularization effects.

\subsubsection{BLEU \& ROUGE Evaluation}
\paragraph{WebGPT.}
\begin{figure}[ht]
\vskip 0.2in
\begin{center}
\centerline{\includegraphics[width=\columnwidth]{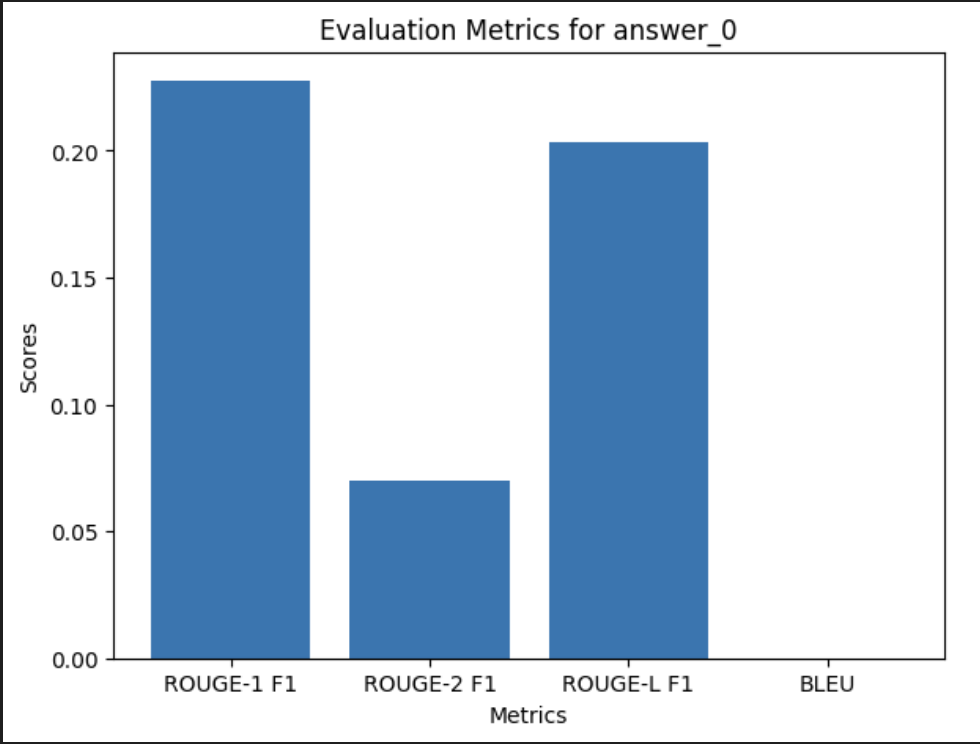}}
\caption{eval answer 0 (Trained)}
\label{fig:webgpt-trained}
\end{center}
\vskip -0.2in
\end{figure}

\begin{figure}[ht]
\vskip 0.2in
\begin{center}
\centerline{\includegraphics[width=\columnwidth]{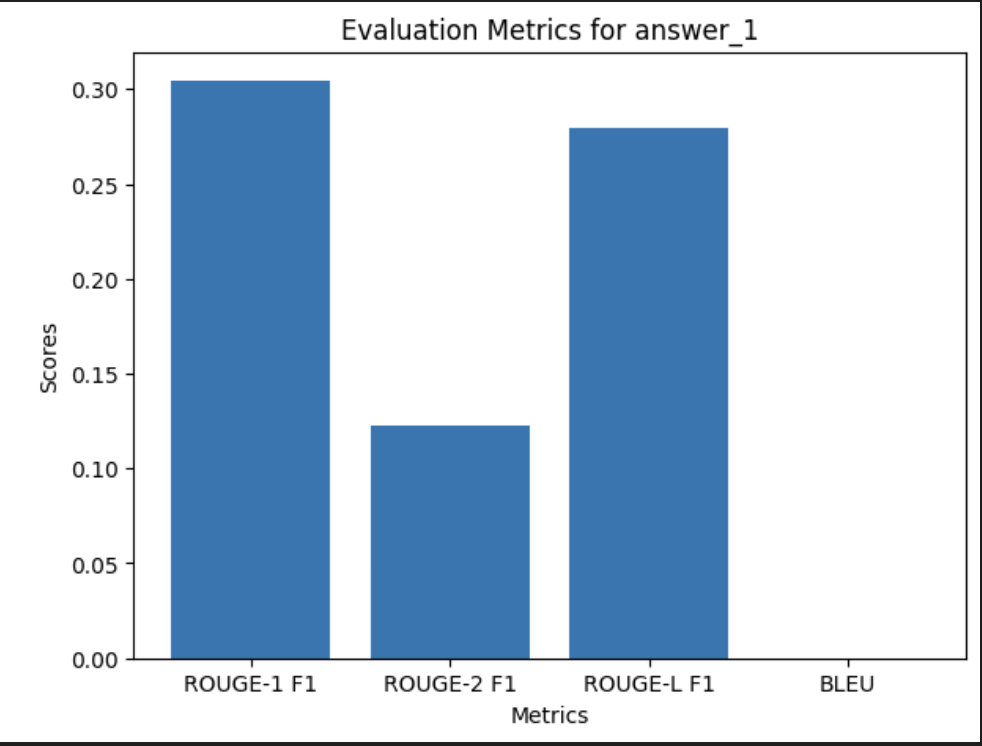}}
\caption{eval answer 1 (Pre-Trained)}
\label{fig:webgpt-pretrained}
\end{center}
\vskip -0.2in
\end{figure}

As the table below shows: the BLEU scores for both models are extremely low, but the trained model has a slightly higher BLEU score compared to the pre-trained model. This indicates that while both models struggle to produce n-grams that match the reference exactly, the trained model performs marginally better.

The pre-trained model outperforms the trained model across all ROUGE metrics (recall, precision, and F1 score). This suggests that the pre-trained model is better at capturing the n-grams present in the reference text, both in terms of individual words (ROUGE-1), bigrams (ROUGE-2), and the longest common subsequences (ROUGE-L). The trained model's lower ROUGE scores indicate that its generated text has less overlap with the reference text compared to the pre-trained model.

\begin{table}[ht]
\centering
\begin{adjustbox}{width=\columnwidth}
\begin{tabular}{@{}lcc@{}}
\toprule
Metric & \textbf{Trained Model (answer\_0)} & \textbf{Pre-trained Model (answer\_1)} \\ \midrule
\textbf{BLEU Score} & $5.317 \times 10^{-60}$ & $5.613 \times 10^{-74}$ \\ \midrule
\textbf{ROUGE-1 Recall} & 0.3054 & 0.4514 \\
\textbf{ROUGE-1 Precision} & 0.2352 & 0.2919 \\
\textbf{ROUGE-1 F1 Score} & 0.2275 & 0.3043 \\ \midrule
\textbf{ROUGE-2 Recall} & 0.1127 & 0.2341 \\
\textbf{ROUGE-2 Precision} & 0.0655 & 0.1144 \\
\textbf{ROUGE-2 F1 Score} & 0.0702 & 0.1230 \\ \midrule
\textbf{ROUGE-L Recall} & 0.2731 & 0.4218 \\
\textbf{ROUGE-L Precision} & 0.2113 & 0.2675 \\
\textbf{ROUGE-L F1 Score} & 0.2035 & 0.2799 \\ \bottomrule
\end{tabular}
\end{adjustbox}
\caption{BLEU and ROUGE Evaluation Results for Trained and Pre-trained Models}
\label{table:evaluation_results}
\end{table}

Overall, The pre-trained model seems to retain better performance in terms of generating text that closely matches the reference in terms of n-gram overlap.Despite the trained model showing a slight improvement in BLEU score, its overall performance, as indicated by the ROUGE metrics, suggests it may not have generalized well from the training data or may require further fine-tuning.

\subsubsection{Cosine-similarity Evaluation}
\paragraph{WebGPT.}
We computed the average cosine similarity across the 100 prompts, resulting in scores of 0.960 for both the post-trained and pre-trained models. This indicates that both models perform similarly well in responding to the prompts from the WebGPT dataset.

Two potential explanations can account for these findings. Firstly, the high cosine similarity score of 0.96 achieved by the pre-trained model suggests that it already delivers excellent results, leaving limited room for improvement in the post-trained model. Secondly, since the majority of the prompts are knowledge-based questions with unique answers, the pre-trained model has demonstrated its proficiency in providing accurate responses to such questions. As a result, the nature of these knowledge-based prompts may not effectively distinguish the performance difference between the post-trained and pre-trained models.

And based on this result, we conclude that the dataset WebGPT is not valid in testing the difference between the post-trained and pre-trained model. And therefore, we try the conversatoinal prompt.

\subsubsection{Human Label Evaluation}
\paragraph{WebGPT.}
We extract some example answers and evaluate the quality of them by ourselves. 
\\\\\textit{\textbf{Example:}}
\\\textit{Prompt:}
The Met Office became part of which government department in July 2011?\\\\\textit{\textbf{Post-trained model answer:}}
The Met Office became part of the Department for Business, Innovation and Skills (BIS) in July 2011. This change was made as part of the government's reorganisation of public sector bodies. The Met Office is now an executive agency of BIS, and is responsible for providing weather and climate science and services.\\\\\textit{\textbf{Pretrained model answer:}}
The Met Office, the United Kingdom's national weather service, was established in 1854. It is responsible for providing weather forecasts and severe weather warnings for the UK and its overseas territories.\\\\In example one, both models provided detailed and accurate answers. In this instance, the DPO training did not enhance the performance beyond that of the pre-trained model, as the latter already achieved satisfactory results.\\\\Upon manual review of all the answers provided in the first 100 prompts, we observed that this pattern held true for the majority of the prompts. Even through human judgment, we did not observe a significant enhancement in model performance through DPO training, particularly for knowledge-based questions.

\textbf{Conversational Prompt}:

\textbf{Case Example 1}

Prompt: "Can you tell me a joke about programming?"

Post-trained model answer: "Why do programmers prefer dark mode? Because the light attracts bugs!"

Pre-trained model answer: "Why don't programmers like nature? It has too many bugs."

In this example, both models provide jokes related to programming. However, the post-trained model's joke is more contextually relevant to a common preference among programmers (dark mode), whereas the pre-trained model's joke is more generic. This indicates the post-trained model's improved ability to tailor responses to specific contexts.

\textbf{Case Example 2}

Prompt: "What advice would you give to someone learning to code?"

Post-trained model answer: "Start with small projects and gradually increase the complexity. Practice consistently, and don't be afraid to ask for help from the community."

Pre-trained model answer: "Learn the basics, practice a lot, and read books about programming."

In this case, the post-trained model offers more detailed and practical advice, including the importance of community support and starting with manageable projects. The pre-trained model's response is accurate but lacks the depth and specificity that makes the post-trained model's advice more actionable and helpful.

These examples highlight the post-trained model's superior performance in providing contextually appropriate and detailed responses compared to the pre-trained model.




\section{Conclusion}

In this project, we explored the application of Direct Preference Optimization (DPO) to fine-tune a pre-trained language model, cognitivecomputations/dolphin-2.1-mistral-7b, leveraging human-labeled preference data. Our goal was to enhance the model's performance on various NLP tasks while simplifying the training pipeline and improving computational efficiency.

Our experimental results demonstrate that DPO can simplify the training process by eliminating the need for reward modeling, thus reducing computational overhead. The DPO algorithm effectively enhances the relative log probability of preferred responses over non-preferred ones, leading to a model that aligns closely with human judgment and expectations.

The evaluation using BLEU, ROUGE, and cosine similarity metrics indicates that while both the trained and pre-trained models perform similarly in many respects, the trained model showed marginal improvements in specific areas. The BLEU scores, although extremely low, suggest that the trained model performs slightly better in terms of n-gram overlap compared to the pre-trained model. However, the pre-trained model outperforms the trained model across all ROUGE metrics, indicating better performance in capturing n-grams and longer text sequences.

Our human evaluations also revealed that the trained model did not significantly outperform the pre-trained model in terms of factual accuracy and conversational style, particularly for knowledge-based questions. This suggests that the pre-trained model already had a high baseline performance, leaving limited room for improvement through DPO.

\section{Discussion}
\subsection{Training Instability and Loss Convergence}
Throughout the training process, we observed instances of instability, particularly the spike in training loss at step 4500. This instability could be attributed to anomalies in the training data or issues with the learning rate. Despite these fluctuations, the overall trend in training loss showed a significant reduction, indicating effective learning. Future work should investigate these spikes further to identify their causes, which may involve checking the training data for outliers or adjusting the learning rate schedule.
\subsection{Performance Metrics Analysis}
The mixed results from the BLEU and ROUGE evaluations highlight the complexity of measuring text generation quality. While BLEU focuses on precision, ROUGE emphasizes recall, and our findings suggest that the trained model's improvements in precision were not sufficient to surpass the pre-trained model's recall capabilities. Additionally, the high cosine similarity scores for both models in the WebGPT dataset indicate that both models are effective in generating semantically similar responses. However, this also suggests that the WebGPT dataset may not be ideal for distinguishing performance differences between the models, particularly for knowledge-based prompts.

In Order for further improvement, it is possible to consider additional fine-tuning with a larger or more diverse dataset to help the model learn better representations and improve its ability to generate text that matches the reference. Or use additional evaluation metrics like METEOR or CIDEr to get a more comprehensive assessment of the model's performance.

\subsection{Human Evaluations}
The human evaluations provided valuable qualitative insights into the models' performance. The lack of significant improvement in the trained model's responses suggests that further fine-tuning or alternative training approaches may be necessary to achieve noticeable enhancements. This also underscores the importance of human evaluations in complementing quantitative metrics to provide a holistic assessment of model performance.

\subsection{Future Work}
Future research should focus on addressing the observed training instability and exploring alternative datasets and evaluation methods to better distinguish the performance of trained models. Exploring hybrid approaches that combine DPO with other reinforcement learning techniques might also yield better results:

\begin{itemize}
    \item \textbf{Model Evaluation and Validation}
    \begin{itemize}
        \item \textit{Cross-Validation}: Use cross-validation to assess the robustness and generalizability of the model.
        \item \textit{Ablation Studies}: Perform ablation studies to understand the impact of different components and configurations on the model's performance.
    \end{itemize}
    \item \textbf{Model Deployment}
    \begin{itemize}
        \item \textit{API Deployment}: Deploy the trained model as an API using frameworks like FastAPI or Flask to make it accessible for real-time applications.
        \item \textit{Edge Deployment}: Optimize and deploy the model on edge devices, ensuring it runs efficiently in resource-constrained environments.
    \end{itemize}
    \item \textbf{Model Enhancement}
    \begin{itemize}
        \item \textit{Ensemble Methods}: Combine multiple models using ensemble techniques to improve performance and robustness.
        \item \textit{Knowledge Distillation}: Use knowledge distillation to create a smaller, faster model that retains the performance of the larger model.
    \end{itemize}
\end{itemize}

\section{Supplemental --Github code}

\href{https://github.com/Yvonneqq9/cs234-chatbot-RLHF}{Our GitHub Repository Link}

\appendix

\end{document}